\documentclass[10pt,twocolumn,letterpaper]{article}

\usepackage{cvpr}
\usepackage{times}
\usepackage{epsfig}
\usepackage{graphicx}
\usepackage{amsmath}
\usepackage{amssymb}
\usepackage{multirow}
\usepackage{bm}
\usepackage{mwe}
\usepackage{pifont}%
\usepackage[abs]{overpic}

\newcommand{\ifod}{iFSD}%
\newcommand{\ifodFull}{Incremental Few-Shot Detection}%
\newcommand{\name}{ONCE}%
\newcommand{\nameFull}{OpeN-ended Centre nEt}%
\newcommand{\keypoint}[1]{\vspace{0.1cm}\noindent\textbf{#1}\quad}
\newcommand{\cut}[1]{}

\usepackage[font=footnotesize,labelfont=bf]{caption}

\usepackage[pagebackref=true,breaklinks=true,letterpaper=true,colorlinks,bookmarks=false]{hyperref}

\cvprfinalcopy %

\ifcvprfinal\fi
\begin{document}

\title{Incremental Few-Shot Object Detection}

\author{
\begin{tabular*}{0.85\linewidth}{@{\extracolsep{\fill}}cccc}
	Juan-Manuel P\'erez-R\'ua$^{1}$ & Xiatian Zhu$^{1}$ & Timothy Hospedales$^{1,3}$ & Tao Xiang$^{1,2}$ \\
	\tt\small j.perez-rua & \tt\small xiatian.zhu & \tt\small t.hospedales & \tt\small tao.xiang \\
	\vspace{-4mm}
	\\
	\multicolumn{4}{c}{\footnotesize *@samsung.com}\\
	\vspace{-4mm}
	\\
	\multicolumn{4}{c}{$^1${ Samsung AI Centre, Cambridge}}\\
	\multicolumn{4}{c}{$^2${ University of Surrey}~~~~~~$^3${ University of Edinburgh}} \\
	\multicolumn{4}{c}{{United Kingdom}}\\
\end{tabular*}
}

\maketitle

\begin{abstract}
Most existing object detection methods rely on the availability of abundant labelled training samples per class and offline model training in a batch mode. 
These requirements substantially limit their scalability to open-ended accommodation of novel classes with limited labelled training data.
We present a study aiming to go beyond these limitations by considering the \ifodFull{} (\ifod{}) problem setting, where new classes must be registered incrementally (without revisiting base classes) and with few examples. 
To this end we propose \nameFull{} (\name{}), a detector designed for incrementally learning to detect novel class objects with few examples. This is achieved by an elegant adaptation of the  CentreNet detector to the few-shot learning scenario, and meta-learning a class-specific code generator model for registering novel classes. \name{} fully respects the incremental learning paradigm, with novel class registration requiring only a single forward pass of few-shot training samples, and no access to base classes -- thus making it suitable for deployment on embedded devices. 
Extensive experiments conducted on both the standard object detection 
and fashion landmark detection tasks show the feasibility of \ifod{} for the first time,
opening an interesting and very important line of research.

\end{abstract}

\section{Introduction}

Despite the success of deep convolutional neural networks (CNNs)  \cite{he2016deep,krizhevsky2012imagenet,simonyan2014very,szegedy2015going}
in object detection \cite{ren2015faster,redmon2018yolov3,liu2016ssd,lin2017focal}, 
most existing models can be only trained offline via a lengthy process of many iterations
in a {\em batch} setting. Under this setting, all the target classes are known, each class has a large number of annotated training samples, and all training images are used for training. This annotation cost and training complexity severely restricts the potential for these methods to grow and accommodate new classes online. Such a missing capability is required in robotics applications \cite{lomonaco2017core50,alabachi2019robotDet}, when the detector is running on embedded devices, or simply to scale up to addressing the long tail of object categories to recognise \cite{openlongtailrecognition}. 
In contrast, humans learn new concepts such as object classes incrementally without forgetting previously learned knowledge \cite{french1999catastrophic}, and often requiring only a few visual examples per class \cite{pinker2003mind,biederman1987concepts30000}.
Motivated by the vision of closing this gap between state-of-the-art object detection and human-level intelligence, a couple of very recent studies \cite{yan2019meta,kang2018few}  proposed methods for few-shot object detector learning. %

Nonetheless, both methods \cite{yan2019meta,kang2018few} are fundamentally {\em unscalable} to real-world deployments in open-ended or robotic learning settings, due to lacking the capability of {\em incremental} learning of novel concepts from a data stream over time.
Specifically, they have to perform a costly training/updating of the detection model using the data of both old (base) and new (novel) classes together, whenever a novel class should be added. Consequently, while they successfully reduce annotation requirements, these models essentially reduce to the conventional batch learning paradigm. This leads to a prohibitively expensive quadratic computation cost in number of categories in an incremental scenario, and also raises issues in data privacy over time \cite{de2019continual,rebuffi2017icarl}. Meanwhile, storage and compute requirements prohibit on-device deployment in robotic scenarios where a robot might want to incrementally register objects encountered in the world for future detection \cite{lomonaco2017core50,alabachi2019robotDet}. 

To overcome the aforementioned limitation,
we study a very practical learning setting -- {\em \ifodFull{}} (\ifod{}). The  
\ifod{} setting is defined by:
{\bf (1)} The detection model can be pre-trained in advance on a set of base classes each with abundant training samples available -- it makes sense to use existing annotated datasets to bootstrap a model \cite{lomonaco2017core50}.
{\bf (2)} Once trained, an \ifod{} model should be capable of deployment to real-world applications where novel classes can be registered at any time using only a few annotated examples. The model should provide good performance for all classes observed so far (i.e., learning without forgetting).
{\bf (3)} The learning of novel classes from an unbounded stream of examples should be feasible in terms of memory footprint, storage, and compute cost. Ideally the model should support deployment on resource-limited devices such as robots and smart phones. 

Conventional object detection methods are unsuited to the proposed setting due to the intrinsic need for batch learning on large datasets as discussed earlier. An obvious idea  is to fine-tune the trained model with novel class training data. However without revisiting old data (batch setting) this causes a dramatic degradation in performance for existing categories due to the {\em catastrophic forgetting} challenge \cite{french1999catastrophic}. The state-of-the-art few-shot object detection methods \cite{yan2019meta,kang2018few} suffer from the same problem too, if denied access to the base (old) class training data and adapted sequentially to novel classes (see the evaluations in Table \ref{tab:coco}).

In this work, as the first step towards the proposed incremental few-shot object detection problem in the context of deep neural networks, we introduce {\em \nameFull{}} (\name{}). The model is built upon the recently proposed  CentreNet~\cite{zhou2019objects}, which was originally designed for conventional batch learning of object detection. 
We take a feature-based knowledge transfer strategy, decomposing CentreNet
into class-generic and class-specific components
for enabling incremental few-shot learning. 
More specifically, \name{} uses the abundant base class training data to first train a class-generic feature extractor. This is followed by meta-learning a class-specific code generator %
with simulated few-shot learning tasks. Once trained, given a handful of images of a novel object class, the meta-trained class code generator elegantly enables the \name{} detector to incrementally learn the novel class in an efficient feed-forward manner during the meta-testing stage (novel class registration).  
This is achieved without requiring access to base class data or iterative updating. 
Compared with \cite{kang2018few,yan2019meta}, \name{} better fits the \ifod{} setting in that its performance is {\em insensitive} to the arrival order and choice of novel classes. This is due to not using softmax-based classification but per-class thresholding in decision making.  Importantly, since each class-specific code is generated independent of other classes, \name{} is intrinsically able to maintain the detection performance of the base classes and any novel classes registered so far.

We make three {\bf contributions} in this work:
{\bf (1)} We investigate the heavily understudied \ifodFull{} problem, which is crucial to many real-world applications. To the best of our knowledge, this is the first attempt to reduce the reliance of a deeply-learned object detector on batch training with large base class datasets during few-shot enrolment of novel classes, unlike recent few-shot detection alternatives  \cite{kang2018few,yan2019meta}.
{\bf (2)} 
We formulate a novel {\em \nameFull{}} (\name{})
by adapting the recent CentreNet detector to the incremental few-shot scenario.
{\bf (3)} 
We perform extensive experiments on both standard object detection (COCO \cite{lin2014microsoft}, PASCAL VOC \cite{everingham2010pascal})
and fashion landmark detection (DeepFashion2 \cite{Ge_2019_CVPR}) tasks.
The results show \name{}'s significant performance advantage  over existing alternatives.

\section{Related Work}

\keypoint{Object detection}
Existing deep object detection models fall generally into two categories:
(1) Two-stage detectors \cite{girshick2014rich,girshick2015fast,ren2015faster,he2015spatial,dai2016r},
(2) One-stage detectors \cite{liu2016ssd,redmon2016you,redmon2018yolov3,lin2017focal,zhou2019objects,zhou2019bottom,law2018cornernet}.
While usually being superior in detection performance, the  two-stage methods are less efficient than the one-stage ones due to the need for object region inference (and subsequent classification from the set of object proposals). Typically, both approaches assume a large set of  training images per class and need to train the detector in an offline batch mode. This restricts their usability and scalability when novel classes must be added on the fly during model deployment. Despite being non-incremental, they can serve as the detection backbone of a few-shot detector. Our \name{} method is based on the one-stage
CentreNet \cite{zhou2019objects} which is chosen because of its efficiency and competitive detection accuracy, as well as the fact that it can be easily decomposed into class-generic and specific parts for adaptation to the \ifodFull{} problem. 

\keypoint{Few-shot learning}
For image recognition, efficiently accommodating novel classes on the fly is widely studied under the name of few-shot learning (FSL) \cite{vinyals2016matching,ravi2016optimization,nichol2018first,finn2017model,snell2017prototypical,sung2018learning,wei2019Close}.
Assuming abundant labelled examples of a set of base classes, FSL methods aim to meta-learn a data-efficient learning strategy that subsequently allows novel classes to be learned from very limited per-class examples.  
A large body of FSL work has investigated how to learn from such scarce data without overfitting
\cite{nichol2018first,santoro2016one,Chen_2019_CVPR,Li_2019_CVPR,Chu_2019_CVPR,Gidaris_2019_CVPR,Sun_2019_CVPR,Zhang_2019_CVPR,Tokmakov_2019_ICCV,Li_2019_ICCV,Ravichandran_2019_ICCV,Dvornik_2019_ICCV,Gidaris_2019_ICCV}. 
Nonetheless, these FSL works usually focus on classification of whole images, or  {\em well cropped} object images. This is much simpler than object detection as object instances do not need to be separated from diverse background clutter, or localised in space and scale.  In this work we extend  few-shot classification to the more challenging object detection task.

\keypoint{Few-shot object detection and beyond}
A few recent works have attempted to exploit few-shot learning techniques for object detection \cite{yan2019meta,kang2018few,Karlinsky_2019_CVPR}. However, these differ significantly from ours in that they consider a {\em non-incremental} learning setting, which restricts dramatically their scalability and applicability in scenarios where access to either large-scale base class data is prohibitive. This is the case for example, due to limited computational resources and/or data privacy issues. We therefore consider the more practical {\em incremental} few-shot learning setting, eliminating the infeasible requirement of repeatedly training the model on the large-scale base class training data\footnote{Some works have studied incremental learning of object detectors~\cite{opelt2006incremental}. However, they do not tackle the few shot regime.}. While this more challenging scenario inevitably lead to inferior performance compared to  {\em non-incremental} learning, as shown in our experiments, it is more representative of natural human learning capabilities and thus provides a good research target with great application potential once solved. 

There are other techniques that aim to minimise the amount of data labelling for object detection such as weakly supervised learning \cite{bilen2016weakly,diba2017weakly}
and zero-shot learning \cite{zhu2019zero,rahman2018zero,bansal2018zero}. 
They assume different forms of training data and prior knowledge, and  
conceptually are complementary to our \ifod{} problem setting. They can thus be  combined when there are multiple input sources available (e.g., unlabelled data or semantic class descriptor).

\section{Methodology}
\keypoint{Problem Definition}
We consider the problem of {\em \ifodFull{} (\ifod{})}: obtaining a learner able to {\em incrementally} recognise {\em novel} classes using only a few labelled examples per class.
We consider two {\em disjoint} object class sets: the {\em base classes} used to bootstrap the system, which are assumed to come with abundant labelled data; and the {\em novel classes} which are sparsely annotated, and to be enrolled incrementally.  That is, in a computationally efficient way, and without revisiting the base class data.

\cut{This learning capability is practically critical since an object detection system will be often confronted with  {\em previously unseen classes} during deployment, and the teaching budget available from the environment would be usually limited. As such, the system has to maximise the use of previously acquired knowledge in a flexible and intelligent manner,  as humans are able to do \cite{pinker2003mind}.}

\subsection{Object Detection Architecture}
To successfully learn object detection from sparsely annotated novel classes, we wish to build upon an effective architecture, and exploit knowledge transfer from base classes already learned by this architecture. 
\cut{Due to the requirement of learning the abundant training data of  base classes for knowledge transfer,  we want to establish the learner on an existing object detection method. This resembles the learning process of human infants
whereby  the necessary elementary knowledge can be acquired for 
future exploitation in few-shot supervision setting.}
However, the selection of the base object detection architecture {\em cannot} be arbitrary given that we need to adapt the detection model to novel classes on-the-fly. For example, while Faster R-CNN \cite{ren2015faster} is a common selection and provides strong performance given large scale annotation, it is less flexible to accommodate novel classes due to a two-stage design and the use of softmax-based classification.

In this work, we build upon the recently developed object detection model 
CentreNet \cite{zhou2019objects} based on several considerations:
{\bf (1)} It is a highly-efficient one-stage object detection pipeline with better speed-accuracy trade-off than alternatives such as SSD \cite{liu2016ssd}, RetinaNet \cite{lin2017focal}, and YOLO \cite{redmon2016you,redmon2018yolov3}.
{\bf (2)} Importantly, it follows a {\em class-specific} modelling paradigm, which allows easy and efficient introduction of novel classes in a plug-in manner.
Indeed, the core feature of CentreNet, the per-class
heatmap-based centroid prediction is naturally appropriate for 
incremental learning as required in our setting.

\begin{figure}[t]
\begin{center}
    \begin{overpic}[width=0.40\textwidth]{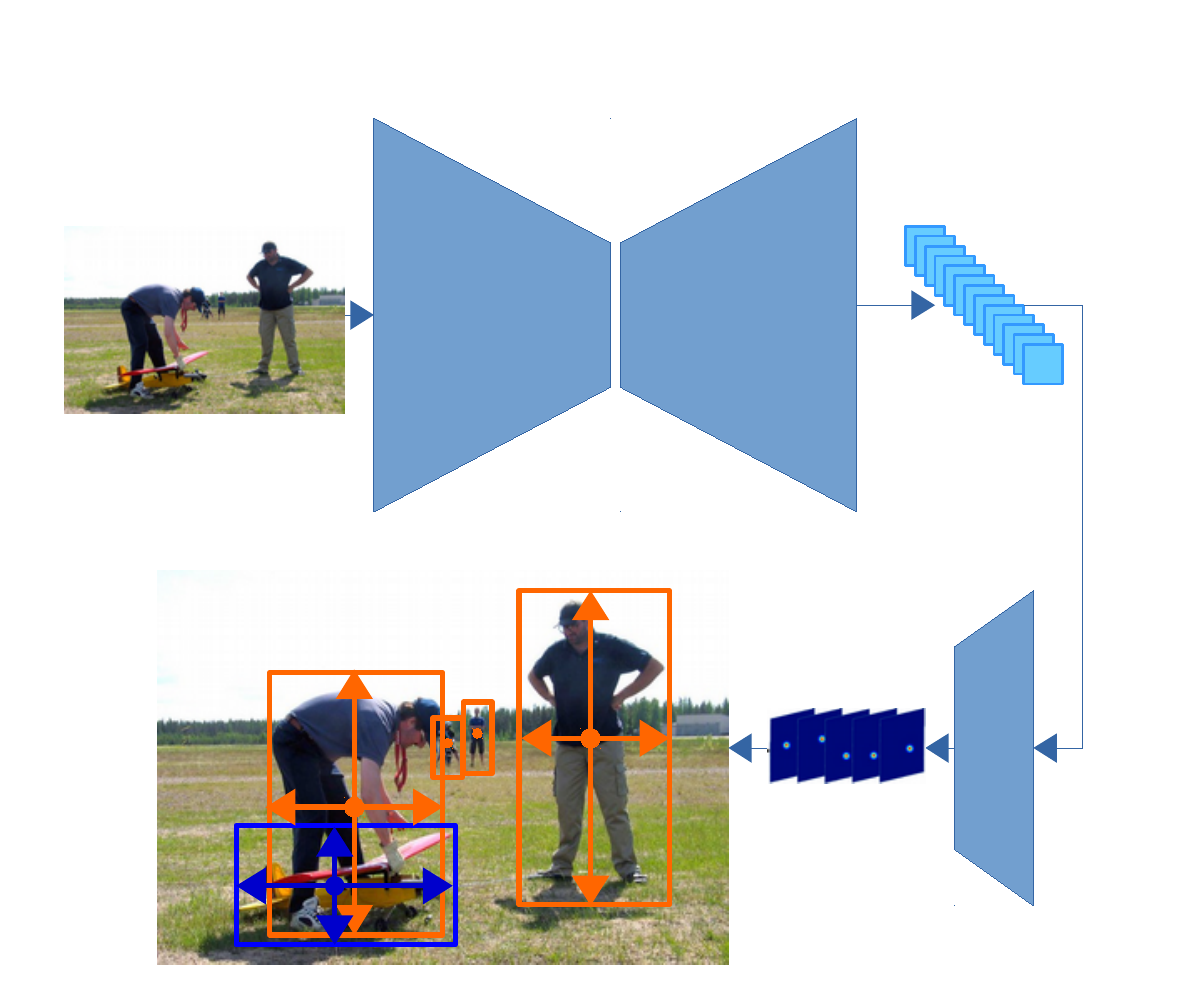}
    \put (18,90) {\scriptsize{Test image~$I$}}
    \put (71.5,80) {\scriptsize{Feature extractor~$f(\cdot)$}}
    \put (152,92) {\tiny{\parbox{1cm}{\centering Feature \\maps~$f(I)$}}}
    \put (162,22) {\rotatebox{90}{\tiny{\parbox{1.2cm}{\centering Object locator~$h(\cdot)$}}}}    
    \end{overpic}
    \caption{
        Overview of Centre-Net. A backbone network, which can
        be implemented with resolution-reducing blocks followed
        by upsampling operations, generates 
        feature maps (top). These feature maps are further 
        processed by a small CNN (the object locator) and transformed into a set of 
        per-class heatmaps encoding the object box centre points and its size (bottom).
    }
    \label{fig:centrenet}
\end{center}
\end{figure}

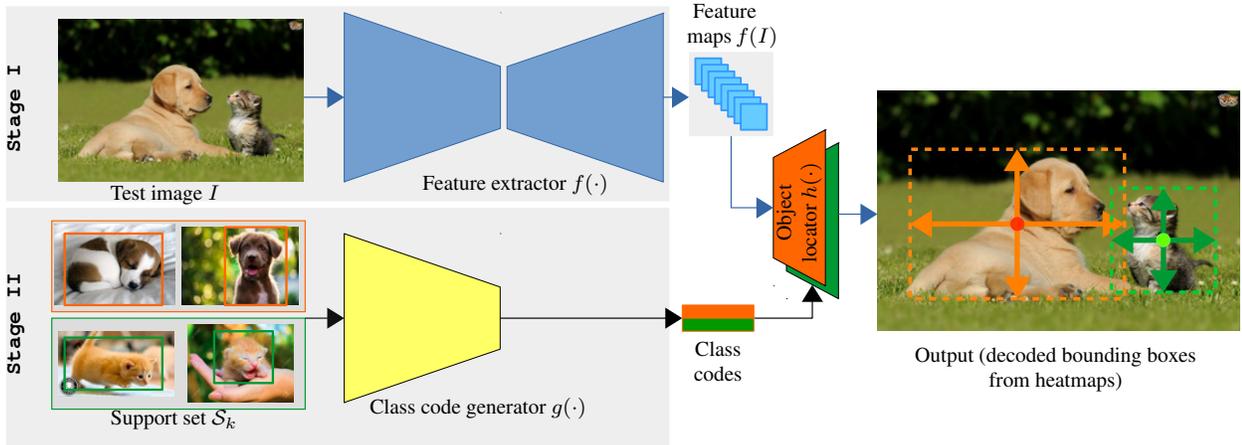
\begin{figure*}[t]
\begin{center}
    \begin{overpic}[width=0.99\textwidth]{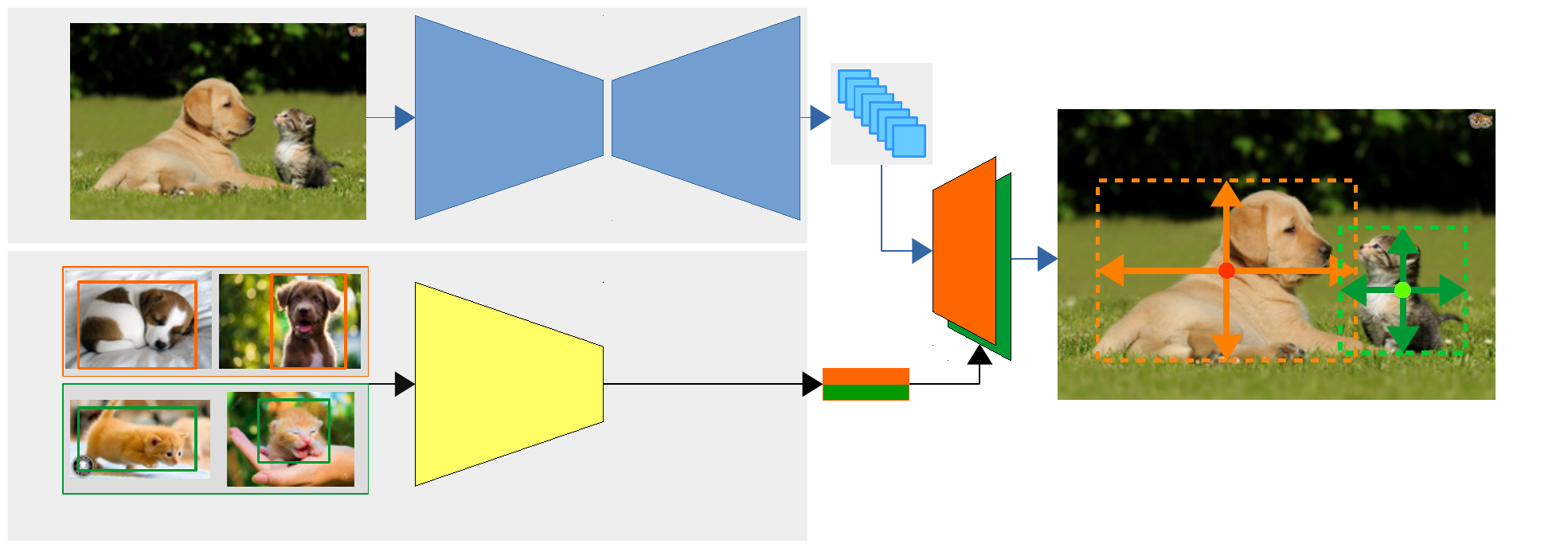}
    \put (3,114) {\rotatebox{90}{\textbf{\footnotesize \texttt{Stage I}}}}
    \put (3,30) {\rotatebox{90}{\textbf{\footnotesize \texttt{Stage II}}}}
    
    \put (42,96) {\footnotesize{Test image~$I$}}
    \put (42,11) {\footnotesize{Support set~$\mathcal{S}_k$}}
    
    \put (160,100) {\footnotesize{Feature extractor~$f(\cdot)$}}
    \put (140,15) {\footnotesize{Class code generator~$g(\cdot)$}}
    
    \put (260,160) {\footnotesize{\parbox{1cm}{\centering Feature \\maps~$f(I)$}}}
    \put (257.5,33) {\footnotesize{\parbox{1cm}{\centering Class codes}}}
    
    \put (294,70) {\rotatebox{90}{\footnotesize{\parbox{1.5cm}{\centering Object locator $h(\cdot)$}}}}
    
    \put (342.5,30) {\footnotesize{\parbox{4cm}{\centering Output (decoded bounding boxes from heatmaps)}}}
    
    \end{overpic}
    \caption{
        The architecture of our {\em \nameFull} (\name) model.
        Specifically, the feature extractor (an encoder-decoder model in our implementation, coloured blue in the top-left) 
        generates the class-generic feature maps $f(I)$ of a test image. These maps
        are further convolved with the class-specific codes (coloured orange for the dog class, and green for the cat class) predicted
        by the class code generator (bottom-left, coloured yellow) from a few labelled support samples per class,
        to generate the object detection result in heatmap format (not shown for simplicity).
        The model training of \name{} involves two stages: 
        {\bf(1)} Stage I: a regular CentreNet-like supervised learning is performed on the
        abundant training data of base classes.
        {\bf(2)} Stage II:  episodic meta-training is performed with the weights of the feature extractor being frozen, allowing the class code generator to learn how to generate a class-specific code from a small per-class support set such that the model can generalise well to unseen novel classes (right).
        The base classes are used as {\em fake} novel classes in meta-training.
        It is noted that \name{} can also be  applied for other detection problems, \eg., fashion landmark localisation.
    }
    \label{fig:pipeline}
\end{center}
\end{figure*}

\subsubsection{A Review of the CentreNet Model}
The key idea of CentreNet is to reformulate object detection as a {\em point+attribute} regression problem.
It is inspired by keypoint detection methods
\cite{newell2016stacked,xiao2018simple},
taking a similar spirit to 
\cite{law2018cornernet,zhou2019bottom}
without the need for point grouping and post-processing.
The architecture of CentreNet is depicted in
Fig.~\ref{fig:centrenet}.
Specifically, as the name suggests, CentreNet takes the centre point and the spatial size (\ie. the width \& height) of an object bounding box as the regression target.
Both are represented with 2D heatmaps, generated
according to the ground-truth annotation. 
In training, the model is optimised to predict such heatmaps, supervised by an $L_1$ regression loss.
For model details, we refer the readers to the original paper due to space limit.

\keypoint{Remarks}
It is worth mentioning that this keypoint estimation based formulation for object detection not only eliminates the need for region proposal generation, but also enables  object location and size prediction in a common format by predicting corresponding pixel-wise aligned heatmaps. For few-shot object detection in particular, a key merit of CentreNet is that each individual class maintains its own prediction heatmap and makes independent detection by activation thresholding. We show next how to exploit this property of CentreNet in order to support incremental enrolment of novel classes in an order-free and combination-insensitive manner, without interference between old \& new classes. This is in contrast to the softmax classification used in existing models \cite{yan2019meta,kang2018few} where interactions between classes make this vision hard to achieve.

\subsection{Incremental Few-shot Object Detection}
As CentreNet is a batch learning model, it is {\em unsuited} for \ifod.
We address this problem by incorporating a
meta-learning strategy \cite{vinyals2016matching,santoro2016one} to CentreNet architecture, 
which results in the proposed \textbf{\em \nameFull} (\name).

\keypoint{Model formulation}
\name{} starts by decomposing CentreNet into two components: 
(i) {\em feature extractor}, %
which is shared by all the base and novel classes,
and (ii) {\em object locator}, %
which contains class-specific parameters for each individual class to be detected. 
Specifically, %
feature extractor takes as input an image,
and outputs
a 3D feature map.
Then, the object locator analyses %
the feature map with a class-specific
code used as convolutional kernel %
and yields the object detection result %
for that class in form of a heatmap.

In a standard extractor/locator decomposition of CentreNet, the object locator still needs to be trained in batch mode, and with large scale training data. In \name{}, we further paramaterise the object locator by a meta-learned generator network, where the generator network synthesises the parameters of the locator network (\ie, the class-specific convolutional kernel weights) given a few-shot support set. In this way, we transform the conventional batch-mode detector learning problem (second CNN in Fig.~\ref{fig:centrenet}, indicated with the tag ``object locator'') into a feed-forward pass of the parameter generator meta-network (class code generator in Fig.~\ref{fig:pipeline}). 
To achieve this, we perform meta-learning to train the class code generator to solve few-shot detector learning tasks via weight synthesis given a support set (resulting in the green and orange class-specific object locators in Fig.~\ref{fig:pipeline}).

\keypoint{Meta-Training: Learning a Few-Shot Detector}
To fully exploit the base classes with rich training data, we train \name{} in two sequential  stages. In the first stage, we train the class-agnostic feature extractor on the base-class training data. This feature extractor is then fixed in subsequent steps as per other few-shot strategies \cite{sung2018learning}.
In the second stage, we learn few-shot object-detection by jointly training the object locator, which is conditioned on a class-specific code; along with a meta-network that generates it given a support set. This is performed by episodic training that simulates few-shot episodes that will be encountered during deployment. In the following sections we describe the training process in more detail.

\keypoint{Meta-Testing: Enrolling New Classes}
At test time, given a support set of novel classes each with a few labelled bounding boxes,
we directly deploy the trained feature extractor, object locator, and code generator learned during meta-training above. The meta-network generates object-specific weights from the support-set (few-shot) and the object locator uses these to detect objects in the test images. This means that novel class objects are detected in test images in a feed forward manner, {\em without} model training and/or adaptation.

\subsubsection{Stage I: Feature Extractor Learning}
We aim to learn a class-agnostic feature extractor $f$ in \name{}. 
This can be simply realised by standard supervised learning with bound-box level supervision on the base classes, similarly to the original CentreNet~\cite{zhou2019objects}.
Concretely, we perform keypoint detection and train the detection model (including both feature extractor $f(\cdot)$ and object locator $h(\cdot)$) by heatmap regression loss. In this stage we train a complete feature extractor and object locator pipeline, although the objective of this stage is solely to learn a robust feature extractor $f(\cdot)$. The locator learned in this stage is a regular CentreNet locator, which will be discarded in stage II, but will be used at the test time for the base classes.

Given a training image $I \in \mathcal{R}^{h\times w \times 3}$
of height $h$ and width $w$,
we extract a class-agnostic feature map $\mathbf{m}=f(I)$, $\mathbf{m} \in \mathcal{R}^{\frac{h}{r}\times \frac{w}{r} \times c}$. The object locator then detects objects of each class $k$ by processing the feature map with a learned class convolutional kernel $\mathbf{c}_k \in \mathcal{R}^{1 \times 1 \times c}$, where $r$ is the output stride and $c$ the number of feature channels.
We then obtain the heatmap prediction $Y_k \in \mathcal{R}^{\frac{h}{r}\times \frac{w}{r}}$ for class $k$ as:
\begin{equation}
Y_k = h(\mathbf{m}, \mathbf{c}_k) =  \mathbf{m} \odot \mathbf{c}_k, \;\; k \in \{1, 2, \cdots, K_b \},
\label{eq:pred}
\end{equation} 
where $\odot$ denotes the convolutional operation,
and $K_b$ denotes the number of base classes.

For locating the object instances of class $k$, 
we start %
by identifying local peaks $\mathcal{P}_k=\{(x_i, y_i) \}_{i=1}^n$,
which are the points whose activation value is higher or equal to 
its 8-connected spatial neighbours in $Y_k$. 
The bounding box prediction is inferred as 
\begin{align}\label{eq:cn_bbox}
    (x_i + \delta x_i - w_i/2, \;\; y_i + \delta y_i - h_i / 2, \\ \nonumber
    x_i + \delta x_i + w_i/2, \;\; y_i + \delta y_i + h_i / 2)
\end{align}
where $(\delta x_i, \delta y_i) = O_{x_i,y_i}$ is the offset prediction, 
$O \in \mathcal{R}^{\frac{h}{r}\times \frac{w}{r} \times 2}$,
and $(h_i,w_i) = S_{x_i,y_i}$ is the size prediction, 
$S \in \mathcal{R}^{\frac{h}{r}\times \frac{w}{r}\times 2}$,
generated by the offset and size codes
in the same fashion as the class codes.
Given the ground-truth bounding boxes
and this prediction, we %
use the $L_1$ regression loss 
for model optimisation on the parameters of feature extractor $f$ and parameters $\mathbf{c}$ of locator $h$.
In practice, the feature extractor model is implemented
with the ResNet-based backbone of~\cite{xiao2018simple}.

\subsubsection{Stage II: Class Code Generator Learning}

The class code parameters $\mathbf{c}$ learned for detection above are fixed parameters for base-classes only. 
To deal with the~\ifod~setting, 
besides these base-class codes we need
an %
 {\em inductive} class code generator $g(\cdot)$ that can efficiently synthesise class codes for novel classes on the fly during deployment, given only a few labelled samples.
To train the class code generator $g(\cdot)$, we exploit an episodic meta-learning strategy \cite{vinyals2016matching}. This uses the base class data to sample a large number of few-shot tasks, thus simulating the test-time requirement of few-shot learning of new tasks. 
While episodic meta-learning is widely used in few-shot recognition, we customise a strategy for {\em detection} here. 

Specifically, we define an \ifod{} task $T$ as uniform distribution over possible class label sets $L$, each with one or a few unique classes.
To form an {\em episode} to compute gradients and train the class code generator $g(\cdot)$, we start by sampling a class label set $L$ from $T$ (\eg, $L = \{person, bottle, \cdots \}$). With $L$, we sample a {\em support} (meta-training) set $\mathcal{S}$ and a {\em query} (meta-validation) set $\mathcal{Q}$.
Both $\mathcal{S}$ and $\mathcal{Q}$ are labelled samples of the classes in $L$.

In the forward pass, the support set $\mathcal{S}$
is used for generating a class code for each sampled class $k$ as: %
\begin{equation}
    \tilde{\bm{c}}_k = g(\mathcal{S}_k), %
    \label{eq:code_gen}
\end{equation}
where $\mathcal{S}_k$ are the support samples of class $k$.
With these codes $\{ \tilde{\bm{c}}_k \}$, our method then performs object detection for query images $I$ by using the feature extractor (Eq.~\eqref{eq:meta_feat}) and object locator (Eq.\eqref{eq:meta_pred}):
\begin{equation}
    \bm{m} = f(I), \;\; \text{with} \;\; I \in  \mathcal{Q},
    \label{eq:meta_feat}
\end{equation}
\begin{equation}
    \tilde{Y} = h(\bm{m}, \tilde{\bm{c}}_k).
    \label{eq:meta_pred}
\end{equation}
\name{} is then trained to minimise the mean prediction error on $\mathcal{Q}$
by updating solely the parameters of the code generator
(cf. Eq. \eqref{eq:code_gen}). %
Same as CentreNet, $L_1$ loss is used as the objective function in this stage, 
defined as $|\tilde{Y} - {Z}|$ 
where ${Z}$ is the ground-truth heatmap.

\subsubsection{Meta Testing: Enrolling New Classes}

Given the feature extractor ($f$, trained in Stage I),
the code generator ($g$, trained in stage II), and
the object locator ($h$, Eq. \eqref{eq:pred}),
At test time, \name{} can efficiently enrol any new class with a few labelled samples
in a feed forward manner without model adaptation and update.

The meta-testing for a novel class is summarised as:
\begin{enumerate}
\itemsep0em 
\item Obtaining its class code with a few-shot labelled set using Eq. \eqref{eq:code_gen};

\item Computing the test image features by using Eq. \eqref{eq:meta_feat};

\item Locating object instances of the novel class by Eq. \eqref{eq:pred};

\item Obtaining all the object candidates using Eq. \eqref{eq:cn_bbox};

\item Finding the heatmap local maxima to output the final detection result of that class.
\end{enumerate}
This process applies for the base classes except that {\em Step 1} is no longer needed since their class codes are already obtained from the training stage I (cf. Eq. \eqref{eq:pred}).
In doing so, we can easily introduce novel classes independently which facilitates
the model \ifod{} deployment.

\subsubsection{Architecture}
For the feature extractor function $f$, we start from a strong and simple baseline architecture~\cite{xiao2018simple} that uses ResNet~\cite{he2016deep} as backbone. This architecture consists of an encoder-decoder pair that first extracts a low resolution 3D map and then expands it by means of learnable upsampling convolutions, outputting high resolution feature maps $f(I)$ for an input image $I$. We leverage the same backbone for the class code generator (without the upsampling operations). Before meta-training (stage II), the class code generator weights are initialised by cloning the weights of the encoder part of the feature extractor. The final convolution outputs are globally pooled to form the class codes $\mathbf{c}_k$, giving a code size of 256\footnote{In practice, a novel class code is $3\times 256$ when accounting for class-specific width and height heatmaps. This is omitted during description of our method for simplicity.}. To handle support sets with variable size, we adopt the invariant set representation of~\cite{Zaheer_2017_NIPS} by average pooling of the class code generator outputs for every image $I_i^{k,s}$ in $\mathcal{S}_k$. The code and trained model will be released.

\section{Experiments}

\begin{figure*}[tbh]
	\centering
	\begin{center}
		\setlength{\tabcolsep}{1pt}
		\renewcommand{\arraystretch}{0.5}
		\begin{tabular}{ccccccc}
			\includegraphics[height=0.10\linewidth]{fig/results/boat_1.png} &			
			\includegraphics[height=0.10\linewidth]{fig/results/bus_2.png} &
			\includegraphics[height=0.10\linewidth]{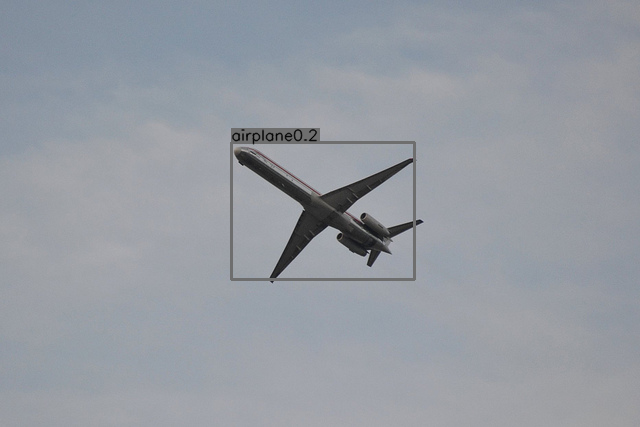} &
			\includegraphics[height=0.10\linewidth]{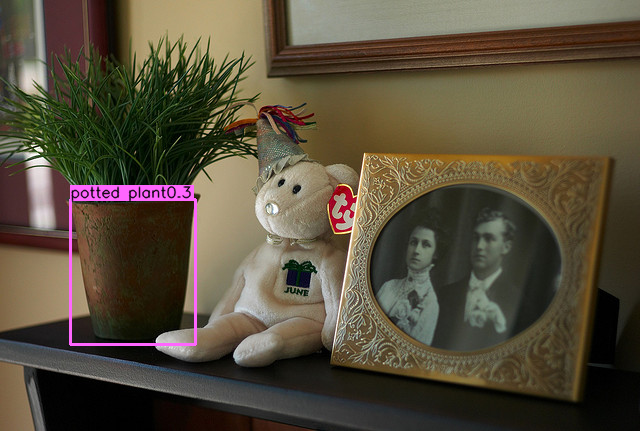} &
			\includegraphics[height=0.10\linewidth]{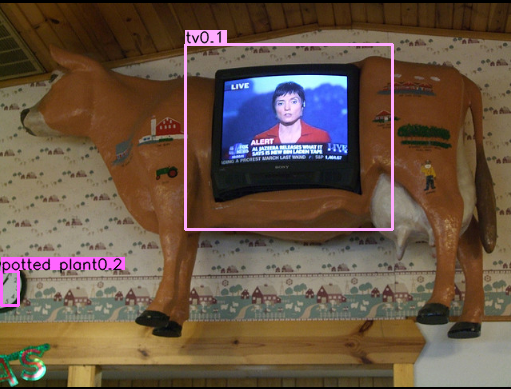} &
			\includegraphics[height=0.10\linewidth]{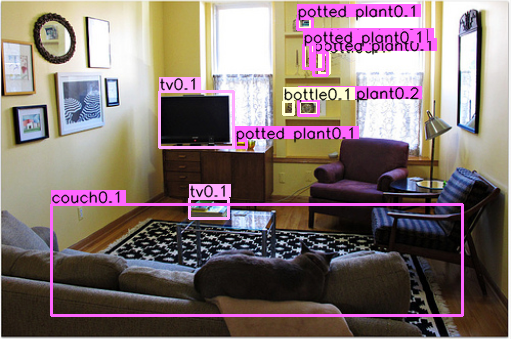} \\
			
			\includegraphics[height=0.10\linewidth]{fig/results/ft_res/arf_boat.png} &
			\includegraphics[height=0.10\linewidth]{fig/results/ft_res/arf_cars.png} &
			\includegraphics[height=0.10\linewidth]{fig/results/ft_res/air_plane.png} &
			\includegraphics[height=0.10\linewidth]{fig/results/ft_res/arf_plant_2.png} &
			\includegraphics[height=0.10\linewidth]{fig/results/ft_res/arf_tv.png} &
			\includegraphics[height=0.10\linewidth]{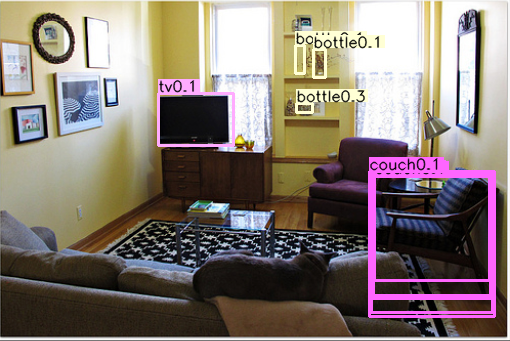} \\
			
		\end{tabular}
	\end{center}
	\caption{Novel class object detection on the COCO {\em val2017} set. %
	Top: our method. Bottom: Fine-Tuning.
	The 10-shot~\ifod{} setting.
	}
	\label{fig:results_coco_10shots}
\end{figure*}

\begin{table}[t]
	\setlength{\tabcolsep}{.1cm}
	\centering
	\footnotesize
	\begin{tabular}{l|cc|cc|cc}
	\hline
	\multirow{2}{*}{Method} 
	& \multicolumn{2}{c|}{Novel Classes}
	& \multicolumn{2}{c|}{Base Classes}
	& \multicolumn{2}{c}{All Classes}
	\\
	& AP & AR
	& AP & AR
	& AP & AR \\
	\hline
	Fine-Tuning %
	&  1.4 & 8.2 & 20.7 & 23.4 & 15.8 & 24.4
	\\
	Feature-Reweight %
	\cite{kang2018few}
	& 5.6 & 10.1 & - & - & - & -
	\\
	Meta R-CNN$^*$ %
	\cite{yan2019meta}  
	&\bf 8.7  &\bf 12.6 & - & - & - & -
	\\
    \bf \name{} %
	& 5.1 & 9.5 & 22.9 & 29.9 & 18.4 & 24.8
    \\
	\hline
	\end{tabular}
	\caption{
	Non-incremental few-shot object detection performance on {\bf COCO  \textit{val2017}} set. 
	Training setting: 10-shot per novel class and all the base class training data.
	`$^\mathbf{*}$': Using different (unknown) support sets of novel classes.
	`-': No reported results.
	}	
	\label{tab:coco_noninc}
\end{table}

\subsection{Non-Incremental Few-Shot Detection}

We start the experimental section with an important contextual experiment.
We evaluated the performance of \name{}
in the {\em non-incremental setting}
as studied in \cite{kang2018few,yan2019meta}.
In particular, we use
COCO \cite{lin2014microsoft}, a popular object detection benchmark, covering 80 object classes from which 20 are left-out to be used as novel classes.
These meta-testing classes happen to be the 20 categories covered by the PASCAL VOC dataset~\cite{everingham2010pascal}.
The remaining 60 classes in COCO serve as base classes. %
For model training, we used 10 shots per novel class along with
all the base class training data.
The results on COCO in Table~\ref{tab:coco_noninc} show that, while not directly comparable due to using different detection backbones
and/or data split, \name{} approaches the performance 
of the two state-of-the-art models \cite{kang2018few,yan2019meta}.
We continue our experimental analysis hereafter with the incremental setting,
which happens to be much more challenging and not
trivially tackled with previous methods~\cite{kang2018few,yan2019meta}.

\subsection{Incremental Few-Shot Object Detection}
\keypoint{Experimental setup}
For evaluating \ifod{}, %
we followed the evaluation
setup of \cite{kang2018few,yan2019meta} but with the key differences that the base class training data is {\em not} accessible during meta-testing, and incremental update for novel classes is required.
In particular, the widely used object detection benchmarks
COCO \cite{lin2014microsoft} and PASCAL VOC %
\cite{everingham2010pascal}  are used. As mentioned before,
 COCO  covers 80 object classes including all 20 classes in PASCAL VOC.
We treated the 20  VOC/COCO shared object classes as novel classes,
and the remaining 60 classes in COCO serve as base classes. This leads to two dataset splits: same-dataset on COCO and cross-dataset with 60 COCO classes as base.

For the \textbf{\em same-dataset} evaluation on COCO,
we used the {\em train} images of base classes 
for model meta-training. In each episode, we randomly sampled 32 tasks, each of which consisted of a 3-class detection problem, and for which each class had 5 annotated bounding boxes per-class.
Larger learning tasks may be beneficial for performance but requiring more GPU memory in training, thus not possible with the resources at our disposal.
For meta-testing, we randomly sampled a support set from the training split of all 20 novel classes. To enrol these 20 novel classes to the model, we consider two settings: {\em incremental batch}, or {\em continuous incremental learning}. In the incremental batch setting, all 20 novel classes are added at once with a single model update. In the continual incremental learning setting, the 20 novel classes are added one by one with 20 models updates.
We evaluated few-shot detection learning with $k \in \{1, 5, 10\}$) bounding boxes annotated for each novel class.
In practice, we used the same support sets of novel classes as \cite{kang2018few}
for enabling a direct comparison between incremental learning and non-incremental learning
(the data split used in \cite{yan2019meta} is not publicly available).
We then evaluated the model performance on the 
{\em validation} set of the novel classes.

For the \textbf{\em cross-dataset} evaluation from COCO to VOC,
we used the same setup and train/test data partitions as above, except that
the model was evaluated on the PASCAL VOC 2007 {\em test} set. That is, the meta-training support/query sets were drawn from COCO, while meta-testing was performed using VOC training data for few-shot detector learning, and VOC testing data for evaluation.

\begin{table*}[ht]
	\setlength{\tabcolsep}{.5cm}
	\centering
	\footnotesize
	\begin{tabular}{c|l|cc|cc|cc}
	\hline
	\multirow{2}{*}{Shot}
	& \multirow{2}{*}{Method} 
	& \multicolumn{2}{c|}{Novel Classes}
	& \multicolumn{2}{c|}{Base Classes}
	& \multicolumn{2}{c}{All Classes}
	\\
	& 
	& AP & AR
	& AP & AR
	& AP & AR \\
	\hline
	\multirow{4}{*}{1} 
	& 
	Fine-Tuning %
	& 0.0 & 0.0 & 1.1 & 1.8 & 0.8 & 1.4\\
    &
	MAML \cite{finn2017model} %
    & 0.1 & 0.5 & N/A & N/A & N/A & N/A \\
    &
    Feature-Reweight$^{\dagger}$ %
	\cite{kang2018few}
    & 0.1 & 0.3 & 2.5 & 4.3 & 1.9 & 3.3 
    \\
	& \bf \name %
	&\bf 0.7 &\bf 6.3 &\bf 17.9 &\bf 19.5 &\bf 13.6 &\bf 16.2\\
	\hline
	\multirow{4}{*}{5} 
	&
	Fine-Tuning %
	& 0.2 & 3.5 & 2.6 & 7.4 & 2.0 & 6.4 \\
    &
	MAML \cite{finn2017model} %
	&  0.4 & 3.9  & N/A & N/A & N/A & N/A\\
	&
	Feature-Reweight$^{\dagger}$ %
	\cite{kang2018few}
    & 0.8 & 5.1 & 3.3 & 8.2 & 2.6 & 7.4
    \\
	& \bf \name %
	&\bf 1.0 &\bf 7.4 &\bf 17.9 &\bf 19.5 &\bf 13.7 &\bf 16.4\\
	\hline
	\multirow{4}{*}{10} 
	&
	Fine-Tuning %
	&  0.6 & 4.2 & 2.8 & 8.0 & 2.3 & 7.0\\
    &
	MAML \cite{finn2017model} %
	&  0.8 & 4.9 & N/A & N/A & N/A & N/A\\
	&
	Feature-Reweight$^{\dagger}$ %
	\cite{kang2018few}
    & \bf 1.5 & \bf 8.3 & 3.7 & 8.9 & 3.1 & 8.7
    \\
	& \bf \name %
	&1.2 & 7.6 &\bf 17.9 &\bf 19.5 &\bf 13.7 &\bf 16.5\\
    \hline
	\end{tabular}
	\caption{Incremental few-shot object detection performance on {\bf COCO {\em val2017}} set.
	{\em Setting}: Incremental learning in batch of all 20 novel classes. 
	`$^{\bm{\dagger}}$':  the code of \cite{kang2018few} is adapted to  use the same detection backbone (CentreNet) and setting for fair comparison.
	}
	\label{tab:coco}
\end{table*}

\keypoint{Competitors}
We compared our \name{} method with several alternatives: (1) The standard {\em Fine-Tuning} method, (2) a popular meta-learning method {\em MAML} (the first-order variant) \cite{finn2017model,nichol2018first}, and (3) a state-of-the-art (non-incremental) few-shot object detection method {\em Feature-Reweight} \cite{kang2018few}.
In particular, since  \cite{kang2018few} was originally designed for the non-incremental setting, we adapted it to the \ifod~setting based on its publicly released code\footnote{The main modification is that only novel classes are visited in meta-test time.}. We note that {\em Meta R-CNN} \cite{yan2019meta} shares the same formulation as \cite{kang2018few}, with the difference of reweighting the regional proposals rather than the whole images. However, without released code we cannot reproduce the Meta R-CNN. %
All methods are implemented on CentreNet/ResNet50 for both the backbone of the detector network and the code generator meta-networks.

\keypoint{Object detection on COCO}
We first evaluated {\em the incremental batch setting}.
The results of all the methods are compared in Table \ref{tab:coco}.
We have several observations:
{\bf(1)} The standard Fine-Tuning method is not only {\em ineffective} for learning from few-shot samples per novel class, but also suffers from catastrophic forgetting (massive base class performance drop), making it unsuited for \ifod.
{\bf(2)} 
As a representative meta-learning method\footnote{Note that there are many more recent FSL methods~\cite{Chen_2019_CVPR,Li_2019_CVPR,Chu_2019_CVPR,Gidaris_2019_CVPR,Sun_2019_CVPR,Zhang_2019_CVPR,Tokmakov_2019_ICCV,Li_2019_ICCV,Ravichandran_2019_ICCV,Dvornik_2019_ICCV,Gidaris_2019_ICCV} that produce better performance on image classification tasks than MAML. However, they are designed for classification only and cannot be easily adapted for the object detection task here. }, MAML improves slightly the few-shot detection capability over Fine-Tuning. 
However, without access to the support sets of base (old) classes in \ifod, 
it is incapable of performing object detection for base classes.
After all, MAML is not designed for incremental learning.
{\bf(3)}
Feature-Reweight, similarly to Fine-Tuning, suffers from catastrophic forgetting when used in the incremental setting. It is inferior to our method for most metrics, 
with a slight edge on novel class detection for the 10-shot experiment. 
This occurs, at the cost of intensive optimisation in testing time, which is not ideal in many practical scenarios.
{\bf(4)}
\name{} achieves the best performance on most experiments for both novel and base classes simultaneously. 
The improvements over baselines are more significant with fewer shots.
In particular, by class-specific detector learning \name{} keeps the performance
on base classes unchanged, naturally solving the learning without forgetting challenge.\footnote{The base class AP we get is lower than in the normal supervised setting.
This is due to early stopping during base class training. Without it, we do reach the results reported in \cite{zhou2019objects}. However, early stopping for training base class detector is important for iFSD as otherwise the features will be overfit to known classes and generalise less to novel classes.} 
{\bf(5)} While the absolute performance on novel classes is still low, this is a new and extremely challenging problem, for which \name{} provides a promising first solution without
requiring test-time optimisation. 
Some qualitative results are shown in Fig.~\ref{fig:results_coco_10shots}.

\begin{figure}[th]
    \centering
    \includegraphics[width=6.4cm]{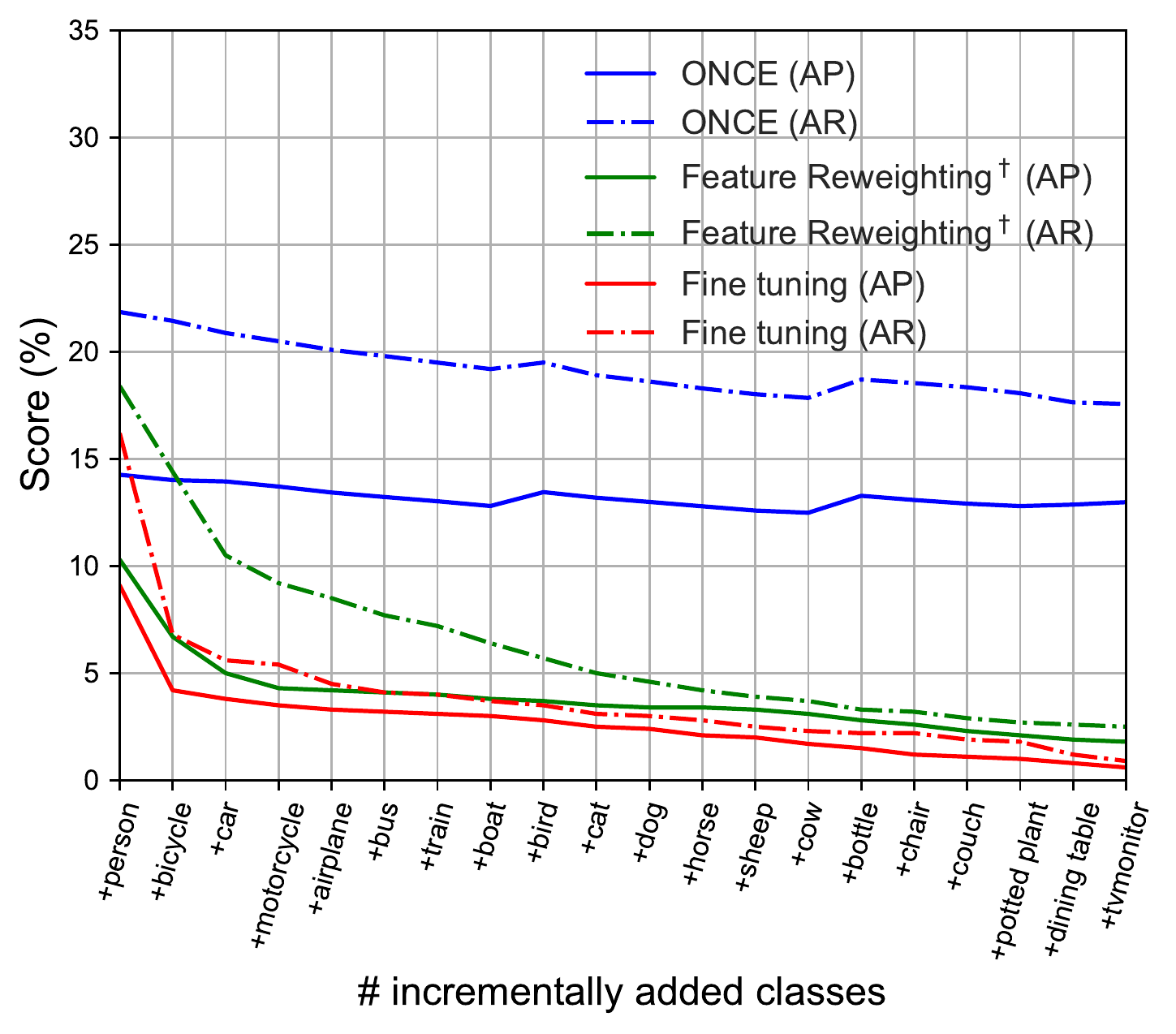}
    \caption{
Incremental few-shot object detection performance on {\bf COCO  {\em val2017}} set. We plot accuracy for all-classes vs. the number of incrementally-added novel classes. Training setting: 10-shot per novel class.
`$^{\bm{\dagger}}$':  the code of \cite{kang2018few} is adapted to  use the same detection backbone (CentreNet) and setting for fair comparison.
            }
    \label{fig:coco_inc}
\end{figure}

We also  evaluated {\em the continuous incremental learning setting}: novel classes are added one at a time. We reported the all-classes accuracy.
In this test, we excluded MAML since it is not designed for \ifod{} with no capability of detecting base class objects. The results in Fig.~\ref{fig:coco_inc} show that the performance of \name{} changes little, while the performance of the competitors drops quickly due to increased forgetting as more classes are added. These results validate our  \name's ability to add new classes on-the-fly. %

\keypoint{Object detection transfer from COCO to VOC. }
We evaluated \ifod{} in a cross-dataset setting from COCO to VOC.
We considered {\em the incremental batch setting},
and reported {\em the novel class performance}
since there are no base class images in VOC. 
The results in Table \ref{tab:pascal}
show that: 
{\bf (1)} The same relative results are obtained as in Table \ref{tab:coco},
confirming that the performance advantages of our model transfers to a test domain different from the training one.
{\bf (2)} Higher performance is obtained on VOC by all methods compared to COCO. This makes sense as COCO images are more challenging and unconstrained than those in VOC.

\begin{table}[t]
	\setlength{\tabcolsep}{2pt}
	\centering
	\footnotesize
	\begin{tabular}{c|l| c | c c c | c | c c c}
	\hline
	Shot & Method 
	& AP & AP$_{S}$ & AP$_{M}$ & AP$_{L}$
	& AR & AR$_{S}$ & AR$_{M}$ & AR$_{L}$
	\\
	\hline
	\multirow{3}{*}{5} 
	&
	Fine-Tuning %
	&  0.1 & 0.1 & 0.8 & 0.3 & 1.9 & 0.7 & 2.9 & 7.6
	\\
    &
	MAML \cite{nichol2018first} %
	& 0.6 & 0.2 & 1.1 & 1.3 & 2.2 & 1.2 & 4.0 & 10.6 \\
	& \bf \name %
	&\bf 2.4 &\bf 1.2 &\bf 2.4 &\bf 3.4 &\bf 12.2 &\bf 5.9 &\bf 16.4 &\bf 33.6
	\\
	\hline
	\multirow{3}{*}{10} 
	&
	Fine-Tuning %
	&  0.3 & 0.1 & 0.8 & 1.0 & 2.8 & 0.9 & 3.3 & 10.2
	\\
    &
	MAML \cite{nichol2018first} %
	& 1.0 & 0.4 & 1.7 & 2.1 & 3.2 & 7.9 & 5.1 & 12.2 \\
	& \bf \name %
	&\bf 2.6 &\bf 5.7 & \bf2.2 &\bf 4.9 &\bf 11.6 &\bf 8.3 &\bf 19.4 &\bf 32.6 \\ 
	\hline
	\end{tabular}
	\caption{Incremental few-shot object detection {\em transfer} performance on {\bf VOC2007 test} set. Training data: COCO. 
	Setting: {\em Incremental batch}.
	We only reported the {\em novel class} performance as there are no base class images in VOC.
}
	\label{tab:pascal}
\end{table}

\subsection{Few-Shot Fashion Landmark Detection}
\keypoint{\bf Experimental setup. }
Besides object detection,
we further evaluated our method for fashion landmark detection 
on the DeepFashion2 benchmark \cite{Ge_2019_CVPR}.
This dataset has 801K clothing items from 13  categories.
A single clothing category contains 8$\sim$19 landmark classes, giving a total of 294 classes.
This forms a two-level hierarchical semantic structure,
which is not presented in the COCO/VOC dataset.
Each image, captured either in commercial shopping stores or in-the-wild consumer scenarios, presents one or multiple clothing items.

On top of the {\em original} train/test data split, 
we developed an \ifod{} setting. Specifically,
we split the 294 landmark classes into three sets: 
153 for training (8 clothing categories), 95 for validation (3 clothing categories), and 46 for testing (2 clothing categories).
This split is category-disjoint across train/val/test sets for testing model generalisation over clothing categories.
The most sparse clothing categories are assigned to the test set.

In each episode of \ifod{} training, we randomly sampled 1 task each with $k$-shot annotated landmarks per class and a total of 5 landmark classes, with $k\in \{1,5,10\}$. 
We used the val set for model selection, \ie. selecting the final model based on the validation accuracy.
To avoid overfitting to the training clothing categories, we randomly sampled 5 out of all the available (8$\sim$19) landmark classes in each  episode.
This is made possible by the class-specific modelling nature of \name, while \cite{kang2018few,yan2019meta} cannot do this.
In meta-testing, we randomly sampled a support set from the {\em original} training set of novel landmark classes (part of the \ifod{} test set),
including $k$ $\in \{1,5,10\}$ bounding boxes annotated for each
novel landmark class, and used it for model learning.
We tested the model performance on the {\em original} testing set of novel landmark 
classes (part of the \ifod{} test set).
We repeated the test process 100 times and report the average.

\keypoint{\bf Competitors. }
In this controlled test, we compared \name{} with the {\em Fine-Tuning} baseline. Other methods for few-shot object detection (\ie., \cite{kang2018few,yan2019meta}) are not trivially adaptable for this task due to its hierarchical semantic structure. Indeed, the inter-class independence 
obtained by adopting CentreNet as detection backbone, and our proposed
few-shot detection framework allows for such general applicability of \name{}.

\keypoint{\bf Evaluation results. }
We evaluated {\em the incremental batch setting}
and reported {\em the novel class performance}.
The results in Table \ref{tab:fashion} show that \name{} consistently and significantly outperforms  {\em Fine-Tuning}. 
This suggests that our model is better at transferring the landmark appearance information
from base classes to novel classes, even when only as little as {\em one shot} training example is available for learning.
due to not having to perform iterative optimisation during meta-test. Note that the absolute accuracy achieved on this task is much higher than the object detection tasks (Table \ref{tab:fashion} vs.~Tables~\ref{tab:coco}\&\ref{tab:pascal}). This is due to the fact that all classes are clothing landmarks so there is much more transferable knowledge from base to novel classes. 
An example of one-shot landmark detection  by \name{} is shown in Fig.~\ref{fig:landmark_exp}. It can be seen that the model can detect landmarks accurately after seeing them only once. 

\begin{figure}[t]
    \centering
    \includegraphics[width=0.825\linewidth]{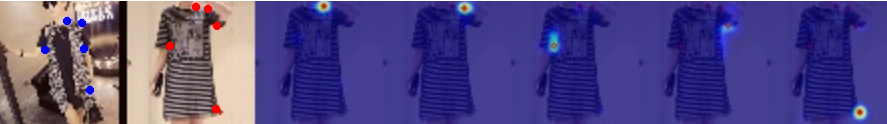}
    \caption{
    Fashion landmark detection by \name. 
    Column {\bf 1}: a support sample with five randomly selected landmark ground-truth.
    Column {\bf 2}: a test image with the predicted landmarks.
    Columns {\bf 3-7}: predicted heatmaps.
            }
    \label{fig:landmark_exp}
\end{figure}

\begin{table}[t]
	\setlength{\tabcolsep}{5pt}
	\centering
	\footnotesize
	\begin{tabular}{c|l| c | c c | c c}
	\hline
	Shot & Method 
	& AP & AP$_{50}$
	& AR & AR$_{50}$
	\\
	\hline
	\multirow{2}{*}{1} &
	Fine-Tuning %
	& 2.8 & 9.0
	& 6.7 & 16.5
	\\
	&\bf \name{} %
	&\bf 4.6 &\bf 26.5
	&\bf 11.8 &\bf 49.9
	\\
	\hline
	\multirow{2}{*}{5} &
	Fine-Tuning %
	& 17.0 & 35.9
	& 25.1 & 42.1
	\\
	&\bf \name{} &\bf 29.5 &\bf 77.5
	&\bf 42.1 &\bf 87.4
	\\
	\hline
	\multirow{2}{*}{10} & 
	Fine-Tuning %
	& 17.1 & 37.1
	& 24.6 & 43.0
	\\
	&\bf \name{} %
	&\bf 32.2 &\bf 79.5
	&\bf 44.3 &\bf 88.3
	\\
	\hline
	\end{tabular}
	\caption{Incremental few-shot landmark detection performance on 
	the novel classes of {\bf DeepFashion2}.
	Setting: {\em Incremental batch}.
	}	
	\label{tab:fashion}
\end{table}

\section{Conclusion}
We have investigated the challenging yet practical 
incremental few-shot object detection problem. 
Our proposed \name{} provides a promising initial solution to this problem. Critically, \name{}  is capable of incrementally registering novel classes with few examples in a feed-forward manner, without revisiting the base class training data. It yields superior performance over a number of alternatives on both object and landmark detection tasks in the incremental few-shot  setting. Our work is evidence of the need for further efforts towards solving more effectively the \ifod{} problem.

{\small
\bibliographystyle{ieee_fullname}
\bibliography{egbib}
}

\end{document}